\documentclass[sigconf]{acmart}
\usepackage{multirow}
\usepackage{color}
\usepackage{graphicx}
\AtBeginDocument{%
  \providecommand\BibTeX{{%
    \normalfont B\kern-0.5em{\scshape i\kern-0.25em b}\kern-0.8em\TeX}}}

\copyrightyear{2023}
\acmYear{2023}
\setcopyright{acmlicensed}\acmConference[MM '23]{Proceedings of the 31st ACM International Conference on Multimedia}{October 29-November 3, 2023}{Ottawa, ON, Canada}
\acmBooktitle{Proceedings of the 31st ACM International Conference on Multimedia (MM '23), October 29-November 3, 2023, Ottawa, ON, Canada}
\acmPrice{15.00}
\acmDOI{10.1145/3581783.3611719}
\acmISBN{979-8-4007-0108-5/23/10}
\acmSubmissionID{135}

\begin{document}

%\title{POAR: Towards Open-World Pedestrian Attribute Recognition}
\title{POAR: Towards Open Vocabulary Pedestrian Attribute Recognition}

\orcid{0000-0003-0179-1396}
\author{Yue~Zhang}
\email{17112065@bjtu.edu.cn}
\affiliation{%
  \institution{College of Computer and Information Engineering and Key Laboratory of Artificial Intelligence and Personalized Learning in Education of Henan Province, Henan Normal University}
  \city{Xinxiang}
  \country{China}
  \postcode{453007}
}

\orcid{0000-0002-4086-0713}
\author{Suchen Wang}
\email{wang.sc@ntu.edu.sg}
\affiliation{%
  \institution{School of Electrical and Electronic Engineering, Nanyang Technological University}
  %\city{Rocquencourt}
  \country{Singapore}
}

\orcid{0000-0003-0097-6196}
\author{Shichao Kan}
%\email{kanshichao@csu.edu.cn
\email{kanshichao10281078@126.com}
\affiliation{%
  \institution{School of Computer Science and Engineering, Central South University}
  \city{Changsha}
  \country{China}}

\orcid{0000-0001-7857-8687}
\author{Zhenyu Weng}
\email{zhenyu.weng@ntu.edu.sg}
\affiliation{%
  \institution{School of Electrical and Electronic Engineering, Nanyang Technological University}
  %\city{Rocquencourt}
  \country{ Singapore}
}

\orcid{0000-0001-6255-9422}
\author{Yigang Cen}
\email{ygcen@bjtu.edu.cn}
\affiliation{%
 \institution{Institute of Information Science and Beijing Key Laboratory of Advanced Information Science and Network Technology, Beijing Jiaotong University}
 \city{Beijing}
 \country{China}}

\orcid{0000-0002-0645-9109}
\author{Yap-peng Tan}
\email{eyptan@ntu.edu.sg}
\affiliation{%
  \institution{School of Electrical and Electronic Engineering, Nanyang Technological University}
  %\city{Rocquencourt}
  \country{Singapore}
}

\renewcommand{\shortauthors}{Yue Zhang et al.}

\begin{abstract}
 
 Pedestrian attribute recognition (PAR) aims to predict the attributes of a target pedestrian. Recent methods often address the PAR problem by training a multi-label classifier with predefined attribute classes, but they can hardly exhaust all possible pedestrian attributes in the real world. To tackle this problem, we propose a novel Pedestrian Open-Attribute Recognition (POAR) approach by formulating the problem as a task of image-text search. Our approach employs a Transformer-based Encoder with a Masking Strategy (TEMS) to focus on the attributes of specific pedestrian parts (e.g., head, upper body, lower body, feet, etc.), and introduces a set of attribute tokens to encode the corresponding attributes into visual embeddings. Each attribute category is described as a natural language sentence and encoded by the text encoder. Then, we compute the similarity between the visual and text embeddings to find the best attribute descriptions for the input images. To handle multiple attributes of a single pedestrian, we propose a Many-To-Many Contrastive (MTMC) loss with masked tokens. In addition, we propose a Grouped Knowledge Distillation (GKD) method to minimize the disparity between visual embeddings and unseen attribute text embeddings. We evaluate our proposed method on three PAR datasets with an open-attribute setting. The results demonstrate the effectiveness of our method as a strong baseline for the POAR task. Our code is available at \url{https://github.com/IvyYZ/POAR}.

 %Instead of learning a specific classifier for each attribute category, our approach models the pedestrian at a part-level and uses a searching method to handle unseen attributes.
\end{abstract}






%\received{21 April 2023}

\maketitle

\section{Introduction}
\label{sec:intro}

\begin{figure}[t]
	\centering
	\includegraphics[width=1.0\linewidth]{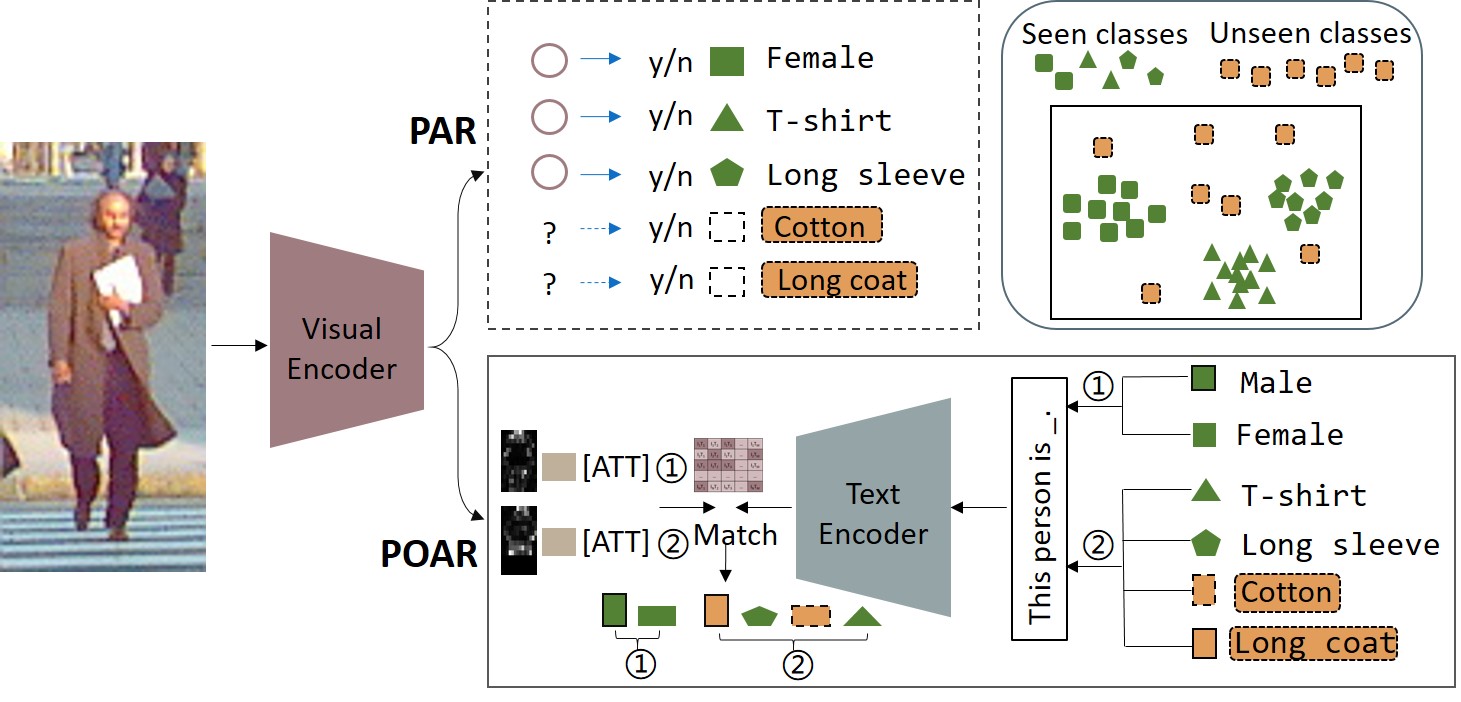}	
	\caption{Comparison of pedestrian attribute recognition (PAR) and pedestrian open-attribute recognition (POAR). The upper part shows the current PAR methods based on multi-label classification, where the attribute categories are predefined. During the test, PAR methods cannot recognize new attributes beyond the predefined classes, such as cotton and long coat. The lower part shows our method, which encodes the images and attributes into a joint image-text feature space. The attribute labels are then determined based on the similarities between the image and attribute embeddings.}
	\label{fig:onecol}
\end{figure}

Pedestrian attribute recognition (PAR) aims to predict attributes of a target pedestrian, such as gender, age, clothing, accessories, etc. Due to the increasing importance of person search~\cite{ZhangPgan,He_2021_ICCV,zhang2022local} and scene understanding~\cite{Wang_2022_CVPR,chang2021comprehensive} in many applications, PAR has emerged as an active research topic in the field of computer vision. Existing methods such as global-local methods~\cite{Tang_2019_ICCV,guo_visual_2022,DBLP:conf/bmvc/LiuLYS18}, attention methods~\cite{liu2017hydraplus,DBLP:conf/bmvc/LiuLYS18}, textual semantic correlations methods~\cite{9782406,2022arXiv220708677L} address the PAR problem by training a multi-label classifier within a predetermined attribute space. Thus they cannot recognize attributes beyond the predefined classes, such as ``cotton'' and ``long coat'' shown in Figure~\ref{fig:onecol}. In this work, we explore how to handle new attributes in an open world and examine a new Pedestrian Open-Attribute Recognition (POAR) problem.

Recently, Radford {\it et al.}~\cite{radford2021learning} proposed a Contrastive Language-Image Pre-training (CLIP) method to model the similarity relationship between images and raw text. CLIP is trained in a task-agnostic setting and can be used to recognize general objects, such as airplane, bird, ocean, and so on. However, it is not yet generalizable on more fine-grained attributes, such as ``upper stride'', ``lower stripe'', ``lower jeans'', in PAR task. For PAR, the challenge is that a pedestrian may have multiple attributes but there are no corresponding location and scale information in the ground truth label set. To address this challenge, part-based~\cite{7139070} and attention-based~\cite{Tang_2019_ICCV} methods have been proposed to localize attributes and learn the attribute-specific features. However, these methods ignore the regional label conflicts, e.g., ``long sleeve'' and ``short sleeve''. Zhao {\it et al.}~\cite{zhao2018grouping} proposed a grouping attribute recognition method by dividing all labels into different groups and localizing regions corresponding to each group attribute based on the detection algorithm. However, the detection algorithm is attribute sensitive. Similarly, we divide the whole attribute classes into multiple groups, and each group corresponds to one visual region, as shown in Figure~\ref{fig:onecol}. But contrastingly, 
we propose a Masking the Irrelevant Patches (MIP) strategy to eliminate insignificant image patches. The appearance of pedestrians can be divided into fixed groups, with certain parts of each group requiring no attention. The proposed strategy does not compromise the ability to identify unseen classes, and it enables accurate localization of regions for improved identification.
%we propose using masks to block out distracting and less informative regions and mask attribute tokens that do not relate to the attribute group. We localize attributes and learn their features based on token prediction with masks rather than detection.

Different from the general PAR task that recognizes only the seen attributes in the training set, our POAR task expands beyond the seen attributes to allow new pedestrian attributes based on the application's needs. To address this challenge, we formulate the POAR problem as an image-text search task, which is trained in a downstream attribute-agnostic manner under the supervision of natural language. Different from CLIP, one image or object has only one class name; our method can use multiple attributes to describe one person in the POAR task. Using Figure~\ref{fig:onecol} as an example, ``male and long coat '' are used to describe the pedestrian. 
To address this problem, we propose multiple attribute tokens ([ATT]) with a masked encoding method in the image encoding step. As shown in Figure~\ref{fig:onecol}, we divide the attributes into multiple groups, and each group is encoded with an attribute token. The masking mechanism is introduced to ensure the attribute tokens are independent from one another. The original image is encoded with multiple attribute tokens corresponding to multiple groups. Based on the masking strategy and the many-to-many property, we propose a Many-To-Many Contrastive (MTMC) loss with masked tokens to guide the update of the network parameter.

We propose an end-to-end, Transformer-based Encoder with a Masking Strategy (TEMS) model. During training, we extract visual and text embeddings using the corresponding encoders. The similarity matrix of image-text pairs in a mini-batch is calculated based on these embeddings. Then, the proposed MTMC loss is used to guide the learning of the model.
During testing, attribute tokens of unseen attributes such as ``cotton'' and ``long coat'' in Figure~\ref{fig:onecol} can be localized based on our MIP strategy. Then, the visual embeddings of unseen attributes are extracted from the localized regions. Meanwhile, text embeddings of these attributes can be extracted using the text encoder. Finally, the attributes of a pedestrian can be recognized based on the similarities of the visual and text embeddings. Considering the potential misalignment between image embeddings and text embeddings of unseen classes, we propose a Grouped Knowledge Distillation (GKD) method by using CLIP model as a teacher network to guide the learning of the embedding space and improve the performance of the proposed TEMS model.
%we implement the knowledge distillation method to decrease the discrepancy between visual embeddings and embeddings of unseen attribute text.
%Because the trained attribute encoder can be used to encode attributes that have not been seen during training, the proposed method can be used to recognize open attributes of pedestrians in the open world. 

Our contributions can be summarized as follows:

\begin{itemize}
	\item
	We formulate the problem of pedestrian open-attribute recognition (POAR) and develop an effective TEMS method as a strong baseline to address it. To the best of our knowledge, this is the first approach to address the POAR problem.
\end{itemize}

\begin{itemize}
	\item	
	We propose an effective masking mechanism to address the localization and encoding of multiple attributes in the TEMS model. Furthermore, we devise a many-to-many contrastive loss with masked tokens to train the network.
\end{itemize}

\begin{itemize}
	\item	
	We propose a grouped knowledge distillation (GKD) method to minimize the disparity between visual embeddings and unseen attribute text embeddings, so that it can be scaled to address more unseen attributes.
\end{itemize}

\begin{itemize}
	\item	
	We evaluate our proposed TEMS method on benchmark datasets with an open-attribute setting, and demonstrate its effectiveness as a strong baseline.
\end{itemize}

%-------------------------------------------------------------------------

%-------------------------------------------------------------------------
\section{Related Work}
\label{sec:rewo}
\begin{figure*}[t]
	\centering
	\includegraphics[width=0.75\linewidth]{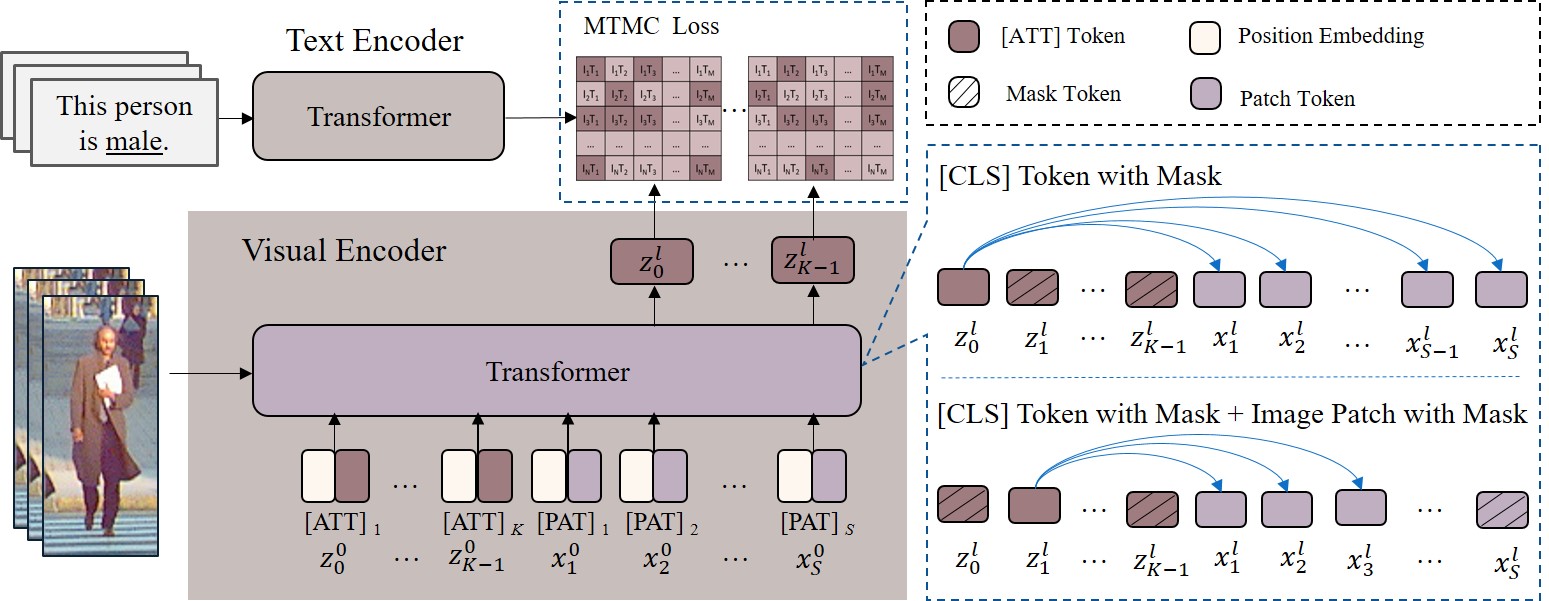}
	\caption{The proposed TEMS framework for POAR task. We evaluate the dot similarity of features from the image and text encoders and then determine the attributes of the pedestrian in the image.}
	\label{fig:framework}
\end{figure*} 

\subsection{Pedestrian Attribute Recognition}
Pedestrian attribute recognition (PAR) has received much interest in person recognition \cite{Cao_2018_CVPR,DBLP:conf/aaai/HandC17,Jia_2021_ICCV,Liu_2016_CVPR,DBLP:journals/corr/abs-2107-12666,wang2022pedestrian} and scene understanding \cite{Jia_2021_ICCV,Liu_2016_CVPR,DBLP:journals/corr/abs-2107-12666}. The mainstream methods address this problem by building a multi-label classifier based on CNN. To improve the recognition accuracy, global methods \cite{li2015multi,liu2017hydraplus}, local methods \cite{Sarafianos_2018_ECCV}, global-local methods \cite{DBLP:conf/bmvc/SarfrazSWS17}, attention-based methods \cite{guo_visual_2022,liu2022dual,lu2023orientation}, sequential prediction methods \cite{he2017adaptively,Yang_2020_CVPR}, curriculum learning methods \cite{zhao2019recurrent}, Graphic model methods \cite{tan2020relation,FanGraph}, and group based methods \cite{zhao2018grouping,jia2022learning} have been proposed. Nikolaos {\it et al.} \cite{Sarafianos_2018_ECCV} proposed an effective method to extract and aggregate visual attention masks across different scales.
Tang {\it et al.} \cite{Tang_2019_ICCV} proposed a flexible attribute localization module to learn attribute-specific regional features. 
These methods focused on domain-specific model designing. To use additional domain-specific guidance, M. Kalayeh {\it et al.} \cite{Kalayeh_2017_CVPR} used semantic segmentation methods to learn attention maps for accurate attribute prediction. 
Liu {\it et al.} \cite{Liu_2016_CVPR} learned clothing attributes with additional landmark labels. 
Yang {\it et al.} \cite{yang2021cascaded} proposed a cascaded split-and-aggregate learning to capture both the individuality and commonality for all attributes.
Li {\it et al.} \cite{2022arXiv220708677L} proposed an image-conditioned masked language model to learn complex sample-level attribute correlations from the perspective of language modeling.
Tang {\it et al.} \cite{tang2022drformer} and Weng {\it et al.} \cite{weng2023exploring} employed ViT as a feature extractor for its nature of modeling long-range relations of regions. Cheng {\it et al.} \cite{9782406} proposed an additional textual modality to explore the textual semantic correlations from attribute annotations. 
These methods are trained on a predefined attribute set and used to recognize the same attributes, which limits the attribute capacity of these models. In our work, we build a TEMS model based on the CLIP \cite{radford2021learning} model to recognize pedestrian open-attributes.
\subsection{Open-Attributes Recognition}
In classification, open-world recognition (OWR) is first proposed by Scheirer \cite{6365193}, which aims to discriminate known from unknown samples as well as classify known ones. Later, prototype-based method \cite{saranrittichai2022multi,vaze2022openset}, knowledge distillation method \cite{gu2021open}, and out of distribution detection method \cite{esmaeilpour2022zero} become popular in image classification and object detection. Definitions and solutions for open-world recognition also differ based on the specific applications.
Esmaeilpour {\it et al.} \cite{esmaeilpour2022zero} used an extended model to generate candidate unknown class names for each test sample and compute a confidence score based on both the known class names and candidate unknown class names for zero-shot out-of-distribution detection. 
Oza {\it et al.} \cite{oza2019c2ae} proposed the class conditional auto-encoder to tackle open-set recognition, which includes closed-set training and open-set training stages. Gu {\it et al.} \cite{gu2021open} adopted an image-text pre-trained model as a teacher model to supervise student detectors. Zhao {\it et al.} \cite{zhao2020object} unified the label space from the training of multiple datasets to improve the generalization ability of a model. In addition, some methods \cite{li2019zero,rahman2020improved} aligned region features and the pre-trained text embeddings in base categories to realize zero-shot detection. Dan {\it et al.} \cite{dan2022enhancing} enhanced class understanding via prompt-tuning for zero-shot text classification. Du {\it et al.} \cite{du2022learning} proposed a detection prompt, to learn continuous prompt representations for open-vocabulary object detection. The CLIP model was proposed by Radford {\it et al.} \cite{radford2021learning}, which performs task-agnostic training via natural language prompting. CLIP can realize zero-shot image recognition. However, CLIP is usually used to recognize general objects, such as airplane, bird, ocean, and so on. For fine-grained attributes recognition, CLIP will fall short in these situations. Our work addresses the pedestrian open-attribute recognition task by identifying both classes, seen and unseen classes, simultaneously.   

\section{Method}
\label{sec:method}
\subsection{Pedestrian Open-Attribute Recognition}

\begin{table*}[htbp]
	\centering
	\caption{Attribute labels of the PETA dataset converted to sequence by prompt.}
	\resizebox{1.7\columnwidth}{!}{%
		\begin{tabular}{lll}
			\hline
			% \multicolumn{3}{c}{PETA dataset} \\
			% \hline
			ATTRIBUTES & KEY   & PROMPT \\
			\hline
			Long, Short  & Hair  & This person has $\left \{ \right\}$ hair. \\
			Male, Female  & Gender & This person is $\left \{ \right\}$. \\
			Less15, Less30, Less45, Less60, Larger60  & Age   & The age of this person is $\left \{ \right\}$ years old. \\
			Backpack, MessengerBag, PlasticBags, Other, Nothing  & Carry & This person is carrying $\left \{ \right\}$. \\
			Sunglasses, Hat, Muffler, Nothing  & Accessory & This person is accessory $\left \{ \right\}$. \\
			LeatherShoes, Sandals, Sneaker, Shoes & Foot  & This person is wearing $\left \{ \right\}$ in foot. \\
			Casual, Formal, Jacket, Logo, ShortSleeve, Plaid, Stripe, Tshirt, VNeck, Other  & Upperbody & This person is wearing $\left \{ \right\}$ in upper body. \\
			Casual, Formal, Trousers, ShortSkirt, Shorts, Plaid, Jeans  & Lowerbody & This person is wearing $\left \{ \right\}$ in lower body. \\
			\hline
	\end{tabular}}%
	\label{tab:prompt}%
\end{table*}%

Pedestrian attribute recognition aims to recognize the fine-grained attributes of a pedestrian (e.g., hairstyle, age, gender, etc.) from the given image. The conventional PAR usually predetermines a set of pedestrian attributes and follows the close-set assumption during the training and test phases. Suppose there are $M$ predetermined pedestrian attributes $\mathcal{A}=\{A_1, A_2, \dots, A_M\}$, e.g., $A_1=$``long hair", $A_2=$``short hair", etc. Then, given a labeled pedestrian dataset $\mathcal{D} = \{(I_i, \mathcal{A}_i)\}_{i=1}^N$, where each image $I_i \in \mathbb{R}^{H\times W\times 3}$ is annotated with a subset $\mathcal{A}_i \subset \mathcal{A}$ of the existing pedestrian attributes. The main objective of PAR is to learn a model to answer which pedestrian attributes from $\mathcal{A}$ appear in a given image $I$.

Existing approaches \cite{Tang_2019_ICCV,guo_visual_2022,9782406} usually convert this to a multi-label classification problem. However, the pedestrian attributes in the real world are potentially unlimited. It is difficult to exhaust all the attributes in a predetermined set and collect the corresponding pedestrian images. Unseen attributes are highly possible to exist in real world applications, while existing classification-based methods \cite{9782406,2022arXiv220708677L} are inherently incapable of handling such cases. To address this limitation, we formulate a pedestrian open-attribute recognition problem in this work. Let $\mathcal{A}_u = \{A_{M+1}, A_{M+2}, \dots, A_{M+M_u} \}$ denote a set of extra attributes that are also of interest in the test phase but not included in the predetermined attribute set $\mathcal{A}$. In POAR, we expect that the model can recognize not only the seen attributes from $\mathcal{A}$ but also the unseen attributes from $\mathcal{A}_u$.

\subsection{Framework}

\begin{figure}[t]
	\centering
	\includegraphics[width=1.0\linewidth]{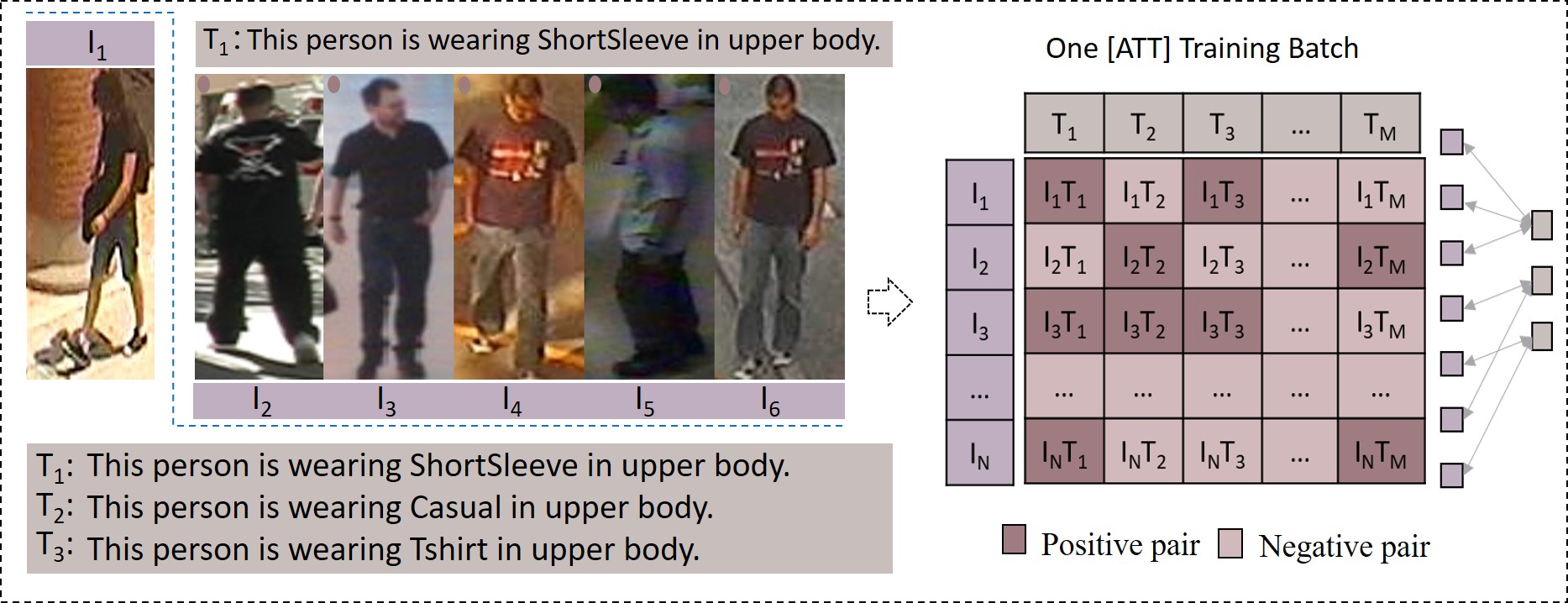}	
	\caption{ Diagram of image-text relationship. The right part represents the many-to-many relationship of image-text.}
	\label{fig:relation}
\end{figure}

\textbf{Overview.}
As illustrated in Figure~\ref{fig:framework}, the proposed Transformer-based encoder with a masking strategy (TEMS) framework consists of an image encoder $\mathbf{\Phi}_I$ and a text encoder $\mathbf{\Phi}_T$. The image encoder processes the input images and derives the visual representation of various pedestrian attributes. Besides, we construct a set of text descriptions (e.g., ``this person has long hair.'', ``this person is carrying backpack.'', etc, and utilize the text encoder of CLIP to encode these attribute descriptions into text embeddings. Then, we compute the vision-text similarity to find the best-matched pedestrian attributes for the input images. Different from the original CLIP, we assume one pedestrian may have more than one associated attribute, as shown in Figure~\ref{fig:relation}. To train our model, we propose a many-to-many contrastive (MTMC) loss with masked tokens to handle the many-to-many relationships between the images and text descriptions. The details of each part are described below.

\textbf{Image Encoding.}
The image encoding process is organized based on token prediction with Transformer. First, the input image $I$ is split into a sequence of non-overlapping small patches \{$P_0$, $P_1$, $\cdots$, $P_{S-1}$\}, where $P_i \in \mathbb{R}^{{r}\times {r}\times 3}$ and $S=\frac{H}{r}\times \frac{W}{r} $. Then, each patch $P_i$ ([PAT]) is projected into the embedding space as a vector $\mathbf{x}_i \in \mathbb{R}^{D}$, where $D$ denotes the feature dimension. Thus, we can obtain $\mathbf{X} = [\mathbf{x}_0, \mathbf{x}_1, \dots, \mathbf{x}_{S-1}] \in \mathbb{R}^{D \times S}$ with $S$ patch embeddings. Inspired by \cite{dosovitskiy2020image,radford2021learning}, we introduce $K$ learnable attribute tokens  $\mathbf{Z} = [\mathbf{z}_0, \mathbf{z}_1, \dots, \mathbf{z}_{K-1}] \in \mathbb{R}^{D\times K}$, where each attribute token $\mathbf{z}_k \in \mathbb{R}^D$ shares the same dimension as $\mathbf{x}_i$.
For each patch embedding and each class token, we generate a corresponding position encoding $\mathbf{e}_j \in  \mathbb{R}^D$, where $j \in \{0, 1, \cdots, S+K-1\}$. The input of the image encoder $\mathbf{\Phi}_I$ is the concatenation of class tokens and patch embeddings combined with their positional embeddings $\mathbf{E}=[\mathbf{e}_0,\mathbf{e}_1,\cdots,\mathbf{e}_{S+K-1}]\in \mathbb{R}^{ D \times (S+K)}$, which is denoted as $\mathbf{V}=[\mathbf{Z},\mathbf{X}]\textcircled{+}\mathbf{E}$, where $\mathbf{V}\in \mathbb{R}^{D\times (S+K)}$, $[,]$ is the concatenate operation, and $\textcircled{+}$ denotes the element-wise summation notation. The output of the image encoder $\mathbf{\Phi}_I$ is the learned embeddings of class tokens, denoted as $\hat{\mathbf{Z}} \in \mathbb{R}^{D\times K}$. The above process can be summarized as follows:
\begin{equation}
\hat{\mathbf{Z}} = \mathbf{\Phi}_I(\mathbf{V}).
\label{image-encoding}
\end{equation}

\textbf{Text Encoding.}
The text encoding is performed by using the text encoder $\mathbf{\Phi}_T$ transferred from the CLIP model. The input of the text encoder is the prompts of pedestrian attributes which are organized as follows. First, the attributes are divided into $K$ groups based on the characteristics of pedestrians, as shown in Table $\ref{tab:prompt}$. Then, we form a prompt template for the attributes of each group. Finally, each prompt sentence in Table $\ref{tab:prompt}$ is tokenized into an embedding using the Byte-Pair-Encoding method\footnote{\url{https://huggingface.co/transformers/v4.8.0/model_doc/clip.html}}. Suppose that the maximum number of words in any of these sentences is $L$. Then, each sentence is tokenized into a vector 
$\mathbf{y}_i \in \mathbb{R}^{L}$. For the $k$-th group, we can obtain $m$ sentence vectors corresponding to the $m$ prompts in this group, denoted as $\mathbf{Y}^k=[\mathbf{y}_1^k,\mathbf{y}_2^k,\cdots,\mathbf{y}_m^k] \in \mathbb{R}^{L\times m}$. The output of the text encoder $\mathbf{\Phi}_T$ is the learned text embeddings $\hat{\textbf{Y}}^k$ for the prompts of the $k$-th group. The above process can be defined as:
\begin{equation}
\hat{\textbf{Y}}^k = \mathbf{\Phi}_T(\mathbf{Y}^k),
\label{text-encoding}
\end{equation}
where $\hat{\textbf{Y}}^k \in \mathbb{R}^{D\times m}$. 

\textbf{Many-to-Many Contrastive Loss.}
For a mini-batch images $\{I_1, I_2, \cdots, I_B\}$, we first extract their token embeddings $\hat{\mathbf{Z}}_b$ and text embeddings $\hat{\mathbf{Y}}_b$ using ($\ref{image-encoding}$) and ($\ref{text-encoding}$), respectively. $\hat{\mathbf{Z}}_b$ and $\hat{\mathbf{Y}}_b$ are sets of token embeddings and text embeddings in all attribute groups related to the given mini-batch images, respectively. Different from the CLIP model, which has a one-to-one relationship between image and text. In the TEMS model, the image and text are involved in a many-to-many relationship, as shown in Figure \ref{fig:relation}. To effectively tackle this scenario, a loss function combined with visual-to-text and text-to-visual contrastive learning is proposed. The visual-to-text contrastive learning term is defined as follows:
\begin{equation}
\mathcal{L}_{v2t}=- {\textstyle \sum_{i=1}^{v}} {\textstyle \sum_{j=1}^{t_i}} log\frac{ exp\left ( \hat{\mathbf{z}}_i^{T} \hat{\mathbf{y}}_{j}^{+}/{\tau }\right )  }{ {\textstyle \sum_{k=1}^{t}}exp\left ( \hat{\mathbf{z}}_i^{T} \hat{\mathbf{y}}_{k}/{\tau }\right )  },
\label{eq:v2t}
\end{equation}
where $\tau$ is a temperature parameter, $t_i$ is the number of positive text embeddings that have the same attribute label with  $\hat{\mathbf{z}}_i$. $v$ and $t$ are the total numbers of token embeddings and text embeddings, respectively. $\hat{\mathbf{z}}_i \in \hat{\mathbf{Z}}_b$, $\hat{\mathbf{y}}_{k} \in \hat{\mathbf{Y}}_b$, $\hat{\mathbf{y}}_{j}^{+} \in \hat{\mathbf{Y}}_b$. $\hat{\mathbf{z}}_i$ and $\hat{\mathbf{y}}_{j}^{+}$ are a positive pair share the same attribute label. Similarly, the text-to-visual contrastive learning term is defined as follows:
\begin{equation}
\mathcal{L}_{t2v}=- {\textstyle \sum_{j=1}^{t}} {\textstyle \sum_{i=1}^{v_j}} log\frac{ exp\left ( \hat{\mathbf{y}}_j^{T} \hat{\mathbf{z}}_{i}^{+}/{\tau }\right )  }{ {\textstyle \sum_{k=1}^{v}}exp\left ( \hat{\mathbf{y}}_j^{T} \hat{\mathbf{z}}_{k}/{\tau }\right )},
\label{eq:t2v}
\end{equation}
where $v_j$ is the number of positive token embeddings that have the same attribute label with  $\hat{\mathbf{y}}_j$. The final loss function is the combination of (\ref{eq:v2t}) and (\ref{eq:t2v}), defined as:
\begin{equation}
\mathcal{L}_{1}=\mathcal{L}_{v2t}+\mathcal{L}_{t2v}.
\label{loss:all}
\end{equation}

%To this end, we have described the main framework of our POAR task. To train the network, we introduce a masking strategy in the following section. 

\subsection{Encoder Networks and Masking Strategy}

\textbf{Image Encoder.}
The image encoder $\mathbf{\Phi}_I$ is a stack of multiple Transformer blocks. Each Transformer block is composed of layer norm (LN) layers  \cite{dosovitskiy2020image}, a multi-head self-attention (MSA) layer  \cite{dosovitskiy2020image}, and a multi-layer perceptron (MLP) network. In the multi-head self-attention layer, the attention weights of each attribute token are automatically calculated among all input tokens. Specifically, the attention weights between the $k$-th attribute token and the others can be computed as
\begin{equation}
\text{Attn}_{\text{token}}(\mathbf{z}_k)=\text{softmax}\Big(\frac{[\mathbf{z}_k^\top \cdot \mathbf{Z}, \;\; \mathbf{z}_k^\top \cdot \mathbf{X}]}{\sqrt{D}}\Big).
\end{equation}

We observe that the first term $\mathbf{z}_k^\top \mathbf{Z}$ often dominates the attention, making the module hardly find the true regions of interest from the input image. This shortcut learning often leads to overfitting and inferior results in our experiments. To address this issue, we mask out the self-attentions between the attribute tokens. This allows useful visual information to be extracted from the input image rather than simply relying on the information that has been extracted by the others. Specifically, we calculate the attention weight of the $k$-th attribute token as
\begin{equation}
\text{Attn}_{\text{mask}}(\mathbf{z}_k)=\text{softmax}\Big(\frac{[\mathbf{z}_k^\top \cdot \mathbf{X}]}{\sqrt{D}}\Big).
\end{equation}

In our experiments, we observe that this technique leads to significant performance gain (Table \ref{tab:abla}). 

In the PAR task, one key challenge is that a pedestrian can have multiple attributes, and there is no corresponding location and scale information in the ground truth label set. We propose to mask the irrelevant patches (MIP) to tackle this challenge. Specifically, we divide the whole attribute classes into multiple groups, and each group ([ATT] token) corresponds to one visual region. Then we mask out regions ([PAT] token) that do not need to pay attention to. For example, the ``hair'' class token pays more attention to the head of the pedestrian, and we block out regions on the lower part of the head region; similarly, the ``upper body'' class token pays more attention to the top part of the image, and we block out the bottom part of the image, etc. 
Thus, the final attention can thus be calculated as follows
\begin{equation}
\text{Attn}_{\text{mask}}(\mathbf{z}_k^l) = \text{softmax}(\frac{\mathbf{z}_k^\top \cdot \mathbf{X} + \mathbf{\varpi }}{\sqrt{D}}),
\end{equation}
where $\mathbf{\varpi } \in \mathbb{R}^{1\times S}$ is a mask vector with value $-\infty$ for the blocked image patches and $0$ otherwise. Taking the ``hair" class token, for example, the region of the head will be $0$, and the remaining regions will be $-\infty$. The output of the self-attention unit is as follows:
\begin{equation}
\mathbf{F}^{l}=\text{Attn}_{\text{mask}}(\mathbf{Z_k}^{l-1})\cdot\mathbf{V}^{l-1},
\end{equation}
where $l$ is the layer index of the self-attention, $\mathbf{Z}^0=\mathbf{Z}$, $\mathbf{V}^0=\mathbf{V}$. The whole process of a Transformer block can be formulated as:
\begin{equation}
{\hat{\mathbf{V}}}^l=\text{MSA}(\text{LN}({\mathbf{V}}^{l-1}))+{\mathbf{V}}^{l-1},
\label{eq:msa}
\end{equation}  
\begin{equation}
{\mathbf{V}}^l=\text{MLP}(\text{LN}({\hat{\mathbf{V}}}^{l}))+{\hat{\mathbf{V}}}^{l}.
\label{eq:mlp}
\end{equation}  

\textbf{Contrastive Learning with Masked Tokens.} 
It should be noted that our contrastive loss is computed based on all the token embeddings and attribute embeddings. The embeddings of those masked tokens are also used to compute the loss. This is because the contrastive loss computed with the masked embeddings will lead to more robust embedding learning.

\subsection{Open-Attribute Recognition}
\begin{figure}[t]
	\centering
	\includegraphics[width=0.85\linewidth]{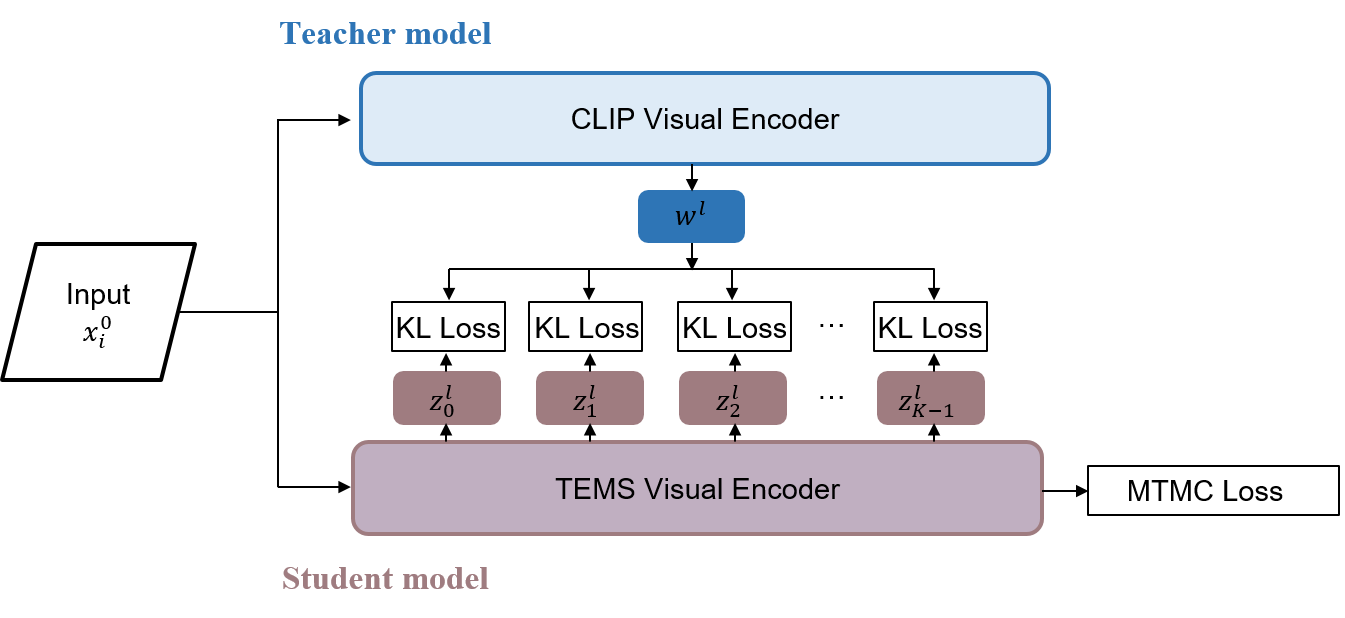}	\caption{The grouped knowledge distillation framework. }
	\label{fig:TS}
\end{figure} 
The knowledge distillation allows for the transfer of knowledge from a complex teacher model to a simpler student model, albeit a slight decrease in performance. The previous paper underscored CLIP's impressive generalization capabilities. Nevertheless, its accuracy in certain tasks, such as pedestrian attribute recognition, leaves room for improvement. To address this, we propose the grouped knowledge distillation (GKD) method, as shown in Figure~\ref{fig:TS}, which leverages the CLIP model as a teacher model and distills its extensive knowledge into our pedestrian attribute model during training. The objective function is defined as:

\begin{equation}
\mathcal{L}_{2} = {\textstyle \sum_{i=1}^{K}} KL(\hat{w}^{l},\hat{z} _{i}^{l}) + \mathcal{L}_{1}.
\label{loss:gkd}
\end{equation}

The Kullback-Leibler (KL) divergence is computed by each visual token embedding $\hat{z} _{i}^{l}$ of the student network and the visual embedding $\hat{w}^{l}$ of the teacher network. These loss functions introduce a non-linear transformation to the input data, which helps to capture more complex and abstract features of the data. The CLIP model has strong generalization performance and can be used as a teacher network to guide the learning of embedding space, reducing semantic disparities between the image embedding features and the unseen text features, and thereby improving the performance of the TEMS model.

%------------------------------------------------------------------------
\section{Experiments}
\begin{table*}[t]
	\centering
	\caption{Performance comparison of POAR experimental results. Blue indicates the model is trained and tested on the same dataset. $*$ denotes our implementation with the official code. }
         \resizebox{1.2\columnwidth}{!}{%
	\begin{tabular}{c|c|cc|cc|cc}
		\toprule
		\multirow{3}[4]{*}{Method} & \multirow{3}[4]{*}{Source Domain} & \multicolumn{6}{c}{Target Domain} \\
		\cmidrule{3-8}          &       & \multicolumn{2}{c|}{PETA} & \multicolumn{2}{c|}{PA100K} & \multicolumn{2}{c}{RAPv1} \\
		&       & R@1   & R@2   & R@1   & R@2   & R@1   & R@2 \\
		\midrule
		CLIP* & –     & 50.2  & 75.7  & 43.4  & 65.9  & 33.6  & 56.5 \\
		%\hline
		VTB*  & PA100K & 31.4  & 62.2  & \textcolor[rgb]{ .357,  .608,  .835}{26.9} & \textcolor[rgb]{ .357,  .608,  .835}{62.2} & 24.2  & 50.7 \\
		\midrule
		TEMS  & \multirow{2}[2]{*}{PETA} & \textcolor[rgb]{ .357,  .608,  .835}{87.6} & \textcolor[rgb]{ .357,  .608,  .835}{96.0} & 45.1  & 73.5  & 42.2 & 68.6 \\
		TEMS+CLIP &       & –     & –     & 44.7  & 74.7 & 42.1  & 69.7 \\
        TEMS (GKD) &       & \textcolor[rgb]{ .357,  .608,  .835}{86.5}    & \textcolor[rgb]{ .357,  .608,  .835}{95.7}     & \textbf{51.9}  & \textbf{78.5} & \textbf{46.2}  & \textbf{71.2} \\
		\midrule
		TEMS  & \multirow{2}[2]{*}{PA100K} & 42.3  & 76.2  & \textcolor[rgb]{ .357,  .608,  .835}{83.3} & \textcolor[rgb]{ .357,  .608,  .835}{92.6} & 39.4  & 63.6 \\
		TEMS+CLIP &       & 50.9  & 77.5  & –     & –     & 39.4  & 64.5 \\
            TEMS (GKD) &       & 50.8  & \textbf{81.4}  &\textcolor[rgb]{ .357,  .608,  .835}{81.3}     & \textcolor[rgb]{ .357,  .608,  .835}{92.8}   & 44.1  & 63.7\\
		\midrule
		TEMS  & \multirow{2}[2]{*}{RAPv1} & 48.8  & 75.0  & 45.1  & 73.1 & \textcolor[rgb]{ .357,  .608,  .835}{80.6} & \textcolor[rgb]{ .357,  .608,  .835}{94.4} \\
		TEMS+CLIP &       & 52.9 & 78.8 & 45.5 & 67.2  & –     & – \\
            TEMS (GKD) &       & \textbf{55.0}  & 76.3   &50.3     & 74.7     & \textcolor[rgb]{ .357,  .608,  .835}{77.5}  & \textcolor[rgb]{ .357,  .608,  .835}{92.3}  \\
		\bottomrule
	\end{tabular}}%
	\label{tab:zero-shot1}%
\end{table*}%

\begin{table}[hbp]
	\centering
	\caption{The number of attribute classes in the test phase.}
        \resizebox{0.8\columnwidth}{!}{%
	\begin{tabular}{c|cc|cc|cc}
		\toprule
		\multirow{3}[4]{*}{Source Domain} & \multicolumn{6}{c}{Target Domain} \\
		\cmidrule{2-7}          & \multicolumn{2}{c|}{PETA} & \multicolumn{2}{c|}{PA100K} & \multicolumn{2}{c}{RAPv1} \\
		& SN    & \multicolumn{1}{l|}{UN} & SN    & UN    & SN    & UN \\
		\midrule
		PETA  & 35    & 0     & 8    & 18    & 12   & 39 \\
		PA100K & 8    & 27    & 26    & 0     & 8     & 43 \\
		RAPv1 & 12    & 23    & 8    & 18    & 51    & 0 \\
		\bottomrule
	\end{tabular}}%
	\label{tab:number}%
\end{table}%
%-------------------------------------------------------------------------
\subsection{Datasets and Experimental Settings}
%analysis.jpg
\begin{figure*}[ht]
	\centering
	\includegraphics[width=1.0\linewidth]{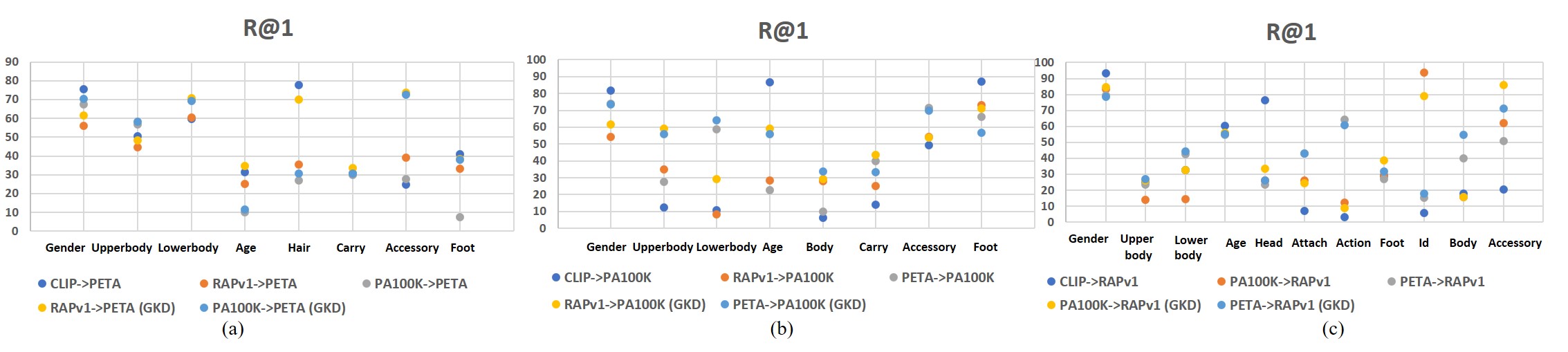}
	\caption{Performance comparison of different groups ([ATT] tokens) by different methods based on the same experimental settings. (a), (b), and (c) represent the results of PETA, PA100K, and RAPv1 datasets, respectively.}
	\label{fig:analysis}
\end{figure*} 
\textbf{Datasets and Evaluation Metrics.}
The proposed method is evaluated on three benchmark datasets, i.e., PETA \cite{deng2014pedestrian}, RAPv1 \cite{li2015multi}, and PA100K \cite{liu2017hydraplus} with an open-attribute setting, meaning the model is trained on one dataset and evaluated on the other two datasets. The classes included in different datasets may not be identical. Recall@K and mA (mean Accuracy) are used to evaluate the performance of our open-attribute recognition model. For the PETA dataset, the ``upper body'' and ``lower body'' groups may include two attribute classes. We select R@2 texts to show for attribute groups in ``upper body'' and ``lower body''.

\textbf{Implementation Details.} 
Our experiments are conducted on the ViT-B/16 backbone networks, which are stacked with 12 Transformer blocks. The input image size is set to 224$\times$224. The value of $r$ is 16, and $K$ is set to be 8, 8, and 11 in PETA, PA100K, and RAPv1, respectively. The learning rate is $5e^{-2}$ with a weight decay of 0.2. The temperature $\tau$ in contrastive loss is set as 5. Data augmentation with random horizontal flip and random erasing are used during training. 

\subsection{Performance Comparison of POAR}
We compare the performance of our method with the CLIP \cite{radford2021learning} and VTB \cite{9782406} methods based on the image-to-text $K$-nearest neighbor retrieval idea, the top-1 and top-2 recall rates are shown in Table~\ref{tab:zero-shot1}. In Table~\ref{tab:number}, SN and UN represent seen and unseen attributes in the different test sets, respectively. 

From Table~\ref{tab:zero-shot1}, we can see that the Recall@1 (R@1) scores of our method are 1.7\% and 8.6\% higher than the CLIP method when the model is trained on the PETA dataset and evaluated on the PA100K and RAPv1 datasets, respectively. When the model is trained on the PA100K and evaluated on the PETA dataset, the R@1 score of our method is 7.9\% lower than the CLIP model. When the model is trained on the RAPv1 and evaluated on the PETA, the R@1 scores of our method are 1.4\% lower than the CLIP model. This is likely due to the fact that the CLIP model is trained on 400 million image-text pairs which have a high probability of containing ``age'' and ``hair'' and other attributes; thus, those unseen attributes defined in our experiments can be considered seen in the CLIP model. However, when we fuse the features of our model and the CLIP model, a higher performance for the POAR task can be obtained. When we train the proposed TEMS model with GKD using the CLIP model as the teacher model, the model's transferred performance to other datasets improves significantly. However, the TEMS model's performance on its training dataset decreases as compared to the results without distillation. For instance, when the model is trained on the PETA dataset, its performance on the PETA test set decreases by including distillation. Nevertheless, the model exhibits significant improvements on other datasets.

To analyze the details of the POAR performance, we show the performance comparison of each group of the CLIP model and the proposed TEMS model in Figure~\ref{fig:analysis}. The results on different datasets and attributes vary between the TEMS method and the CLIP method. However, the model trained through GKD exhibits a significant improvement in the accuracy of various attribute tokens, as evidenced by the comparison between the orange (without GKD) and yellow (with GKD) graphs in Figure~\ref{fig:analysis}. 

\begin{figure}[t]
	\centering
	\includegraphics[width=1.0\linewidth]{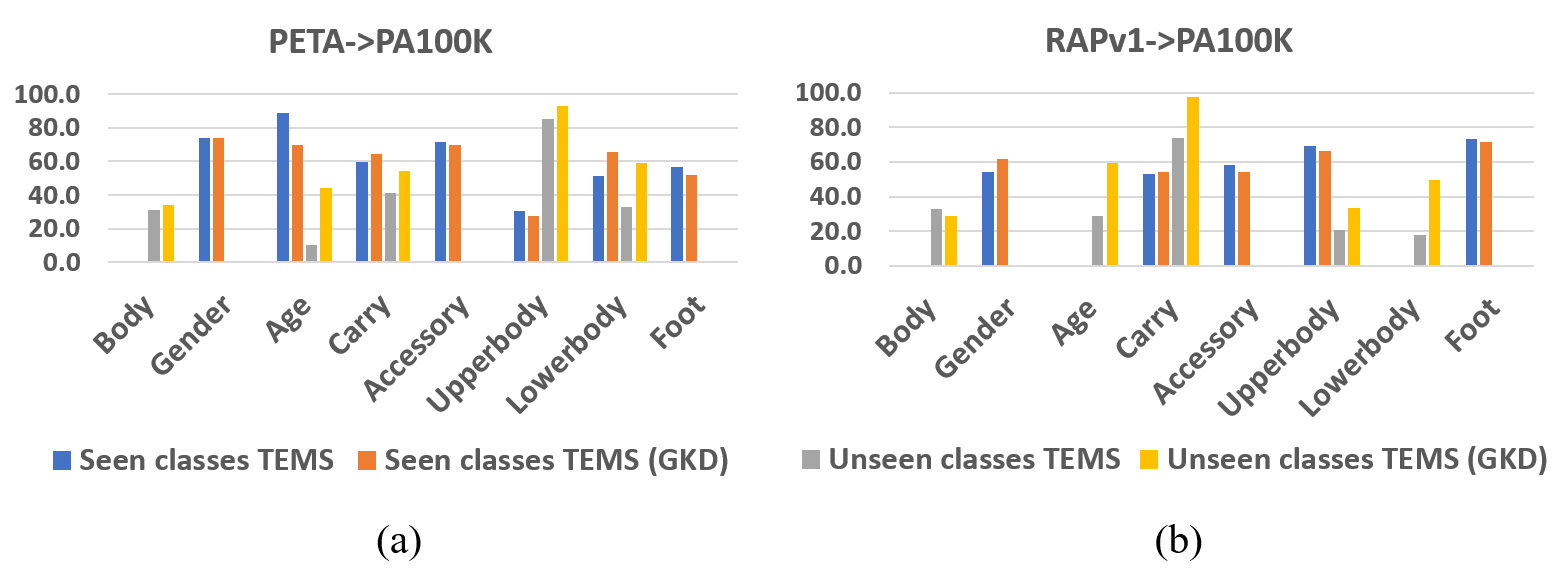}
	\caption{The R@1 scores were calculated for each attribute token in the PA100K dataset, considering both seen and unseen attributes.}
	\label{fig:unseen}
\end{figure} 

\begin{table*}[htbp]
	\centering
	\caption{Results of text-to-image retrieval. Blue indicates that the training and test sets belong to same dataset.}
        \resizebox{1.35\columnwidth}{!}{%
	\begin{tabular}{c|c|ccccccccc}
		\toprule
		& \multirow{3}[6]{*}{Source Domain} & \multicolumn{9}{c}{Target Domain} \\
		\cmidrule{3-11}    Method &       & \multicolumn{3}{c}{PETA} & \multicolumn{3}{c}{PA100K} & \multicolumn{3}{c}{RAPv1} \\
		\cmidrule{3-11}          &       & R@1   & R@5   & R@10  & R@1   & R@5   & R@10  & R@1   & R@5   & R@10 \\
		\midrule
		TEMS & PETA  & \textcolor[rgb]{ 0,  .439,  .753}{90.3} & \textcolor[rgb]{ 0,  .439,  .753}{100.0} & \textcolor[rgb]{ 0,  .439,  .753}{100.0} & 38.5  & 57.7  & 61.5  & 35.3  & 52.9  & 62.8 \\
         %POAR+TS  & PETA  & \textcolor[rgb]{ 0,  .439,  .753}{100.0} & \textcolor[rgb]{ 0,  .439,  .753}{100.0} & \textcolor[rgb]{ 0,  .439,  .753}{100.0} &   &   &   &   &   &  \\
		TEMS  & PA100K & 41.9  & 74.2  & 77.4  & \textcolor[rgb]{ 0,  .439,  .753}{88.5} & \textcolor[rgb]{ 0,  .439,  .753}{96.2} & \textcolor[rgb]{ 0,  .439,  .753}{100.0} & 33.3  & 56.9  & 64.7 \\
         %POAR+TS & PA100K &   &  &   & \textcolor[rgb]{ 0,  .439,  .753}{88.5} & \textcolor[rgb]{ 0,  .439,  .753}{96.2} & \textcolor[rgb]{ 0,  .439,  .753}{100.0} &   &   &  \\
		TEMS  & RAPv1 & 35.5  & 58.1  & 71.0  & 34.6  & 61.5  & 69.2  & \textcolor[rgb]{ 0,  .439,  .753}{96.1} & \textcolor[rgb]{ 0,  .439,  .753}{100.0} & \textcolor[rgb]{ 0,  .439,  .753}{100.0} \\
        %POAR+TS  & RAPv1 &   &  &   &  &   &   & \textcolor[rgb]{ 0,  .439,  .753}{96.1} & \textcolor[rgb]{ 0,  .439,  .753}{100.0} & \textcolor[rgb]{ 0,  .439,  .753}{100.0} \\
		\bottomrule
	\end{tabular}}%
	\label{tab:T2I}%
\end{table*}%

\begin{table}[tbp]
	\centering
	\caption{Evaluation of each component on the PETA dataset. }
	\begin{tabular}{cccc|cccc}
		\toprule
		SAT    & MAT    & MAM    & MIP & mA    & F1    & R@1  & R@2 \\
		\midrule
		$\surd$     &       &       &       & 81.1  & 83.0  & 85.7  & 94.7 \\
		&$\surd$    &       &       & 80.6  & 82.2  & 86.2  & 95.7 \\
		&$\surd$    & $\surd$   &       & 81.0  & 83.0  & 86.4  & 95.7 \\
		& $\surd$    & $\surd$     & $\surd$    & \textbf{83.1} & \textbf{84.4} & \textbf{87.6} & \textbf{96.0} \\
		\bottomrule
	\end{tabular}%
	\label{tab:abla}%
\end{table}%

% Table generated by Excel2LaTeX from sheet 'ablation'
\begin{table}[tbp]
	\centering
	\caption{Our test results for different loss functions on the PETA dataset.  }
        \resizebox{0.7\columnwidth}{!}{%
	\begin{tabular}{cc|cccc}
		\toprule
		OTOC & MTMC & mA    & F1    & R@1 & R@2 \\
		\midrule
		$\surd$   &       & 73.0  & 73.4  & 86.3  & 95.8 \\
		$\surd$     & $\surd$     & 81.5  & 83.3  & 86.3  & 95.8 \\
		&$\surd$    & \textbf{83.1} & \textbf{84.4} & \textbf{87.6} & \textbf{96.0} \\
		\bottomrule
	\end{tabular}}%
	\label{tab:loss}%
\end{table}%

\begin{figure}[t]
	\centering
	\includegraphics[width=1.0\linewidth]{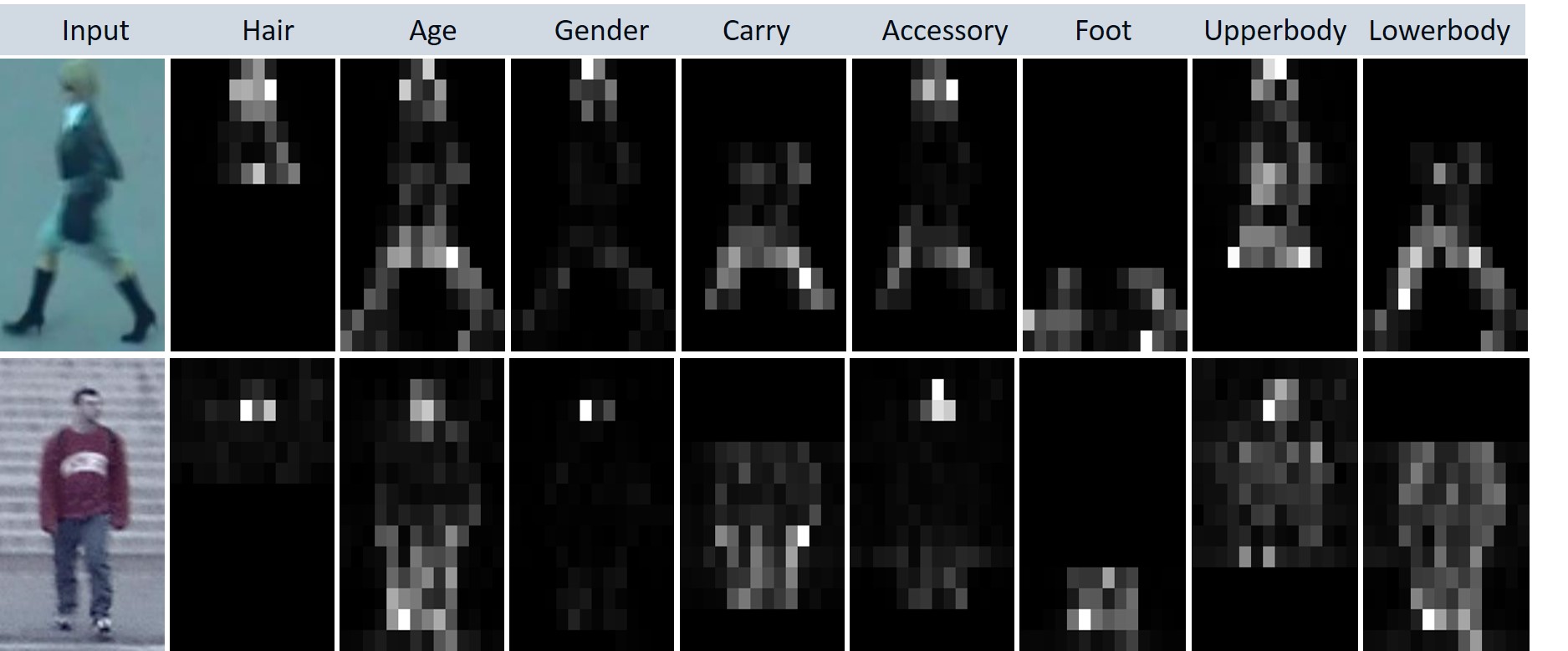}
	\caption{The attention map of each token.}
	\label{fig:attention}
\end{figure} 

% \subsubsection{Attention Map}
\begin{figure*}
	\centering
	\includegraphics[width=0.85\linewidth]{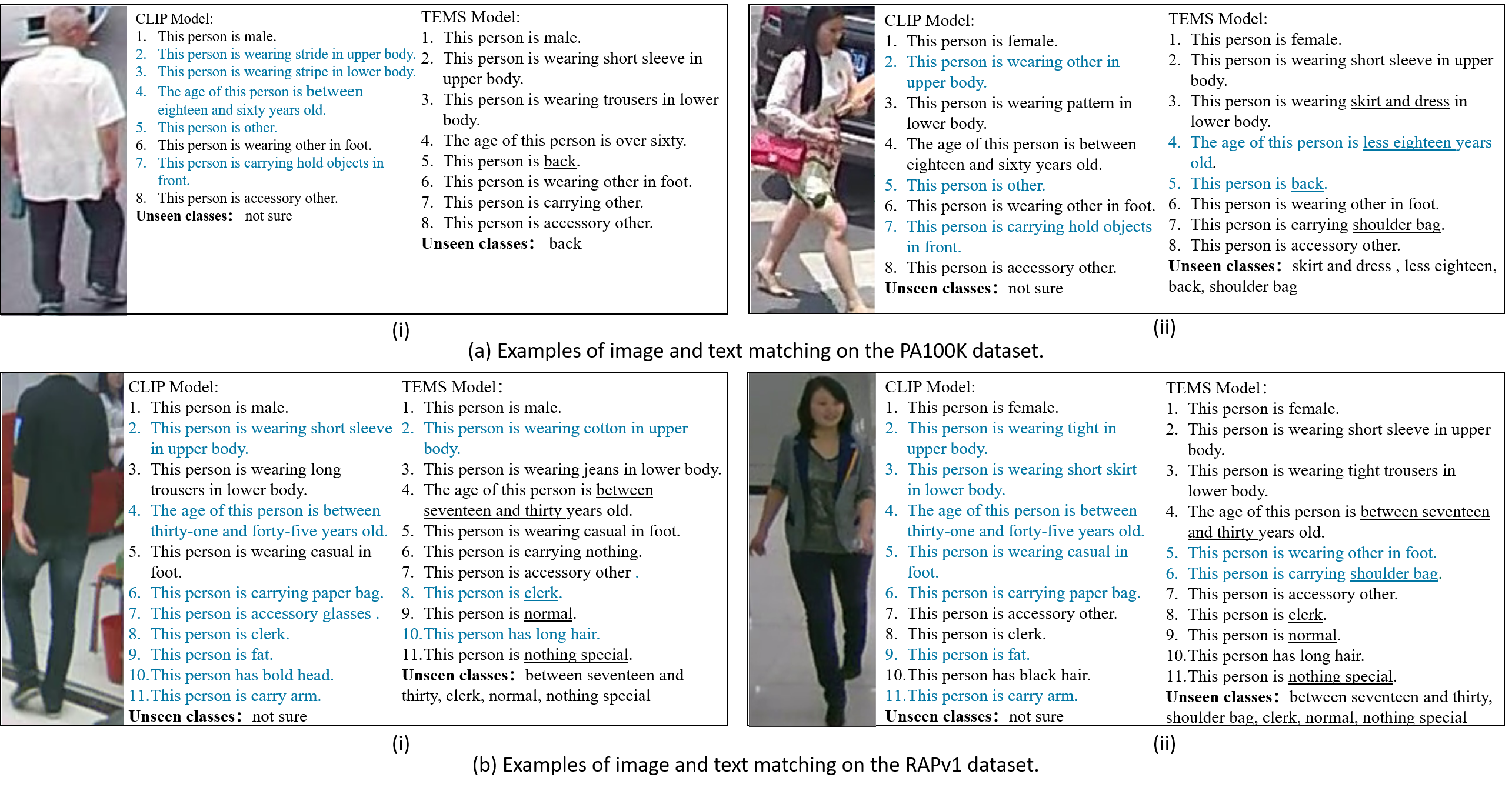}
	\caption{Image-to-text retrieval examples. The model is trained on the PETA dataset. Prompts with blue color indicate that the predicted text is inconsistent with the ground truth. Text with an underline denotes unseen attributes during training.}
	\label{fig:i2t}
\end{figure*}

Furthermore, we compute the recognition performance of seen and unseen attributes in the PA100K dataset, respectively. Results are shown in Figure~\ref{fig:unseen}, the proposed TEMS method has the capability to recognize both seen and unseen categories. For example, when the method is trained on PETA dataset and tested on the PA100K dataset, Figure~\ref{fig:unseen} shows the accuracy of the ``Carry" group for seen (blue) and unseen (gray) classes. Moreover, after implementing GKD training, there is a remarkable enhancement in the model's ability to recognize unseen classes, as evidenced by the comparison of R@1 results for unseen classes in Figure~\ref{fig:unseen} (a) (b) (gray and yellow).
This further confirms the effectiveness of the proposed method in identifying unseen classes.

\subsection{Text-to-Image Retrieval Results}

To examine the generalization ability of our method, we also evaluate the text-to-image retrieval results. The results are reported in Table~\ref{tab:T2I}. Compared to the image-to-text retrieve performance of Table~\ref{tab:zero-shot1}, we can see that the text-to-image recognition task is more challenging than the image-to-text recognition task. This is mainly due to the fact that the text embedding space is more sparse than the image embedding space.

\subsection{Ablation Study}

Our ablation study is conducted on the PETA dataset using the close-set and open-set evaluation mechanisms. Table~\ref{tab:abla} shows the image-to-text retrieval performance of different components in our proposed method, SAT represents a single attribute token. MAT represents multiple attribute tokens. MAM represents multiple attribute tokens with the masking. Table~\ref{tab:loss} shows the image-to-text retrieval performance for different loss functions, OTOC represents one-to-one contrastive loss. MTMC represents many-to-many contrastive loss. 
The one-to-one loss function is performed by defining all attributes of one image in a paragraph description as text input. From Table~\ref{tab:abla}, we can see that each component of the proposed method can contribute positively to the final performance gain. From Table~\ref{tab:loss}, we can see that the many-to-many loss function significantly improves the final performance.

\subsection{Visualization of the Attention Maps}

To illustrate the effectiveness of the proposed masking method, we show the attention map of each attribute token on the PETA dataset in Figure~\ref{fig:attention}. In our method, we proposed a masking strategy to block out distract regions and mask the corresponding attribute tokens. From the attention map, we can see that each attribute token is responsible for a specific part of the body, which shows the effectiveness of the proposed masking method.

\subsection{Image-to-Text Retrieval Examples}

In Figure~\ref{fig:i2t}, we visualize some image-to-text retrieval examples based on embeddings obtained using the CLIP model and our proposed TEMS model. The TEMS model was trained on the PETA dataset. Figure~\ref{fig:i2t} (a) and Figure~\ref{fig:i2t} (b) show examples tested on the PA100K and RAPv1 datasets, respectively. We can see that there are a lot of unseen attributes in the test set, and our model can effectively recognize the unseen attributes. From Figure~\ref{fig:i2t}, the CLIP model exhibits a higher rate of errors in recognizing pedestrian attribute categories. In contrast, as illustrated in Figure ~\ref{fig:i2t} (a) (i), our proposed model achieves perfect recognition of the pedestrian attributes, while the CLIP model only correctly identifies three attributes for the same pedestrian. In the RAPv1 dataset, there are 39 unseen classes, the recognition accuracy for these classes is much higher by the TEMS model than that of the CLIP model. 
 %Despite the presence of numerous unseen categories in the RAP v1 dataset, our TEMS model can accurately identify them. Figure~\ref{fig:i2t} (b) (ii) shows that the proposed model is capable of accurately recognizing pedestrian attributes such as age and occupation.

\section{Conclusions}

In this work, we propose a novel method to tackle the problem of pedestrian open-attribute recognition. Our key idea is to formulate the POAR problem as an image-text search problem. Specifically, we propose a TEMS method to encode image patches with attribute tokens. Then, a many-to-many contrastive loss function with masked tokens is developed to train the model. Finally, the knowledge distillation method is employed to improve the recognition of unseen attribution classes. Experimental results on benchmark PAR datasets with an open-attribute setting show the effectiveness of the proposed method. 
Our proposed POAR task is also promising, as its performance can be further improved by leveraging on the advances in more sophisticated multimodality technologies, like ChatGPT / T5-11B.

\textbf{Limitations:} One limitation of the work is that the text encoder is directly transferred from the CLIP. It is beneficial to examine whether a more effective attribute encoder can be developed in our future work. Another limitation is that the input of the framework is the detected pedestrian; future work could be focused on integrating pedestrian detection and POAR into a unified framework to improve the efficiency of pedestrian open-attribute identification.

\begin{acks}
This work was supported in part by the National Key R\&D Program of China 2021YFE0110500, in part by the National Natural Science Foundation of China under Grant 62202499 and 62062021, in part by the Hunan Provincial Natural Science Foundation of China under Grant 2022JJ40632, in part by the Guiyang scientific plan project contract [2023] 48-11. Corresponding author: Yigang Cen.
\end{acks}

\bibliographystyle{ACM-Reference-Format}
\bibliography{sample-base}

%\end{document}
%\endinput
%%
%% End of file `sample-sigconf.tex'.

\clearpage
\appendix
	
	\section{Supplementary Material}
	
	\subsection{Datasets and Metrics} 
	We evaluate the proposed TEMS baseline on three large-scale benchmark pedestrian attribute datasets.
	
	\textbf{PETA dataset.} Following the standard settings, 9,500 images are used for training, 1,900 images are used for verification, and 7,600 images are used for testing. The model is evaluated on 35 attributes. 
	
	\textbf{RAPv1 dataset.} It is collected from 26 real indoor surveillance cameras and contains 41,585 pedestrians, where 33,268 images are used for training, and 8,317 images are used for testing. The model is evaluated on 51 attributes. 
	
	\textbf{PA100K dataset.} It includes 100,000 images, of which 80,000 images are used for training, 10,000 images are used for validation, and 10,000 images are used for testing. Each image is annotated with 26 commonly used attributes. 
	%The attribute groups of the PA100K and RAPv1 datasets are shown in Table\ref{tab:Gpa100k} and Table\ref{tab:Grapv1}.
	
	\textbf{Metrics.} Our model is trained on one dataset and evaluated on the other two datasets. The top-k results are used to evaluate the performance of our open-attribute recognition model.
	
	\subsection{Ablation Experiment}
	
	Our ablation study is conducted on the RAPv1 dataset using the open-set evaluation mechanisms. Table~\ref{tab:abla} shows the image-to-text retrieval performance of different components in our proposed method. From Table~\ref{tab:abla}, we can see that each component of the proposed method can contribute positively to the final performance gain.
	\begin{table}[htbp]
		\centering
		\caption{Evaluation of each component on the RAPv1 dataset, the model is trained on the PETA dataset. SAT represents a single attribute token. MAT represents multiple attribute tokens. MAM represents multiple attribute tokens with the masking. }
		\begin{tabular}{cccc|cccc}
			\toprule
			SAT & MAT & MAM & MIP & R@1  & R@2 \\ \hline
			$\surd$  & ~ & ~ & ~ & 39.9 & 56.1 \\ \hline
			~ & $\surd$  & ~ & ~ & 41.3 & 61.6 \\ \hline
			~ & $\surd$  & $\surd$  & ~ & 41.1 & 65.7 \\ \hline
			~ & $\surd$  & $\surd$  & $\surd$  & 42.2 & 68.6 \\ 
			\bottomrule
		\end{tabular}%
		\label{tab:abla}%
	\end{table}%
	
	\subsection{Text-to-image Retrieval Examples}
	\begin{figure}[t]
		\centering
		\includegraphics[width=1.05\linewidth]{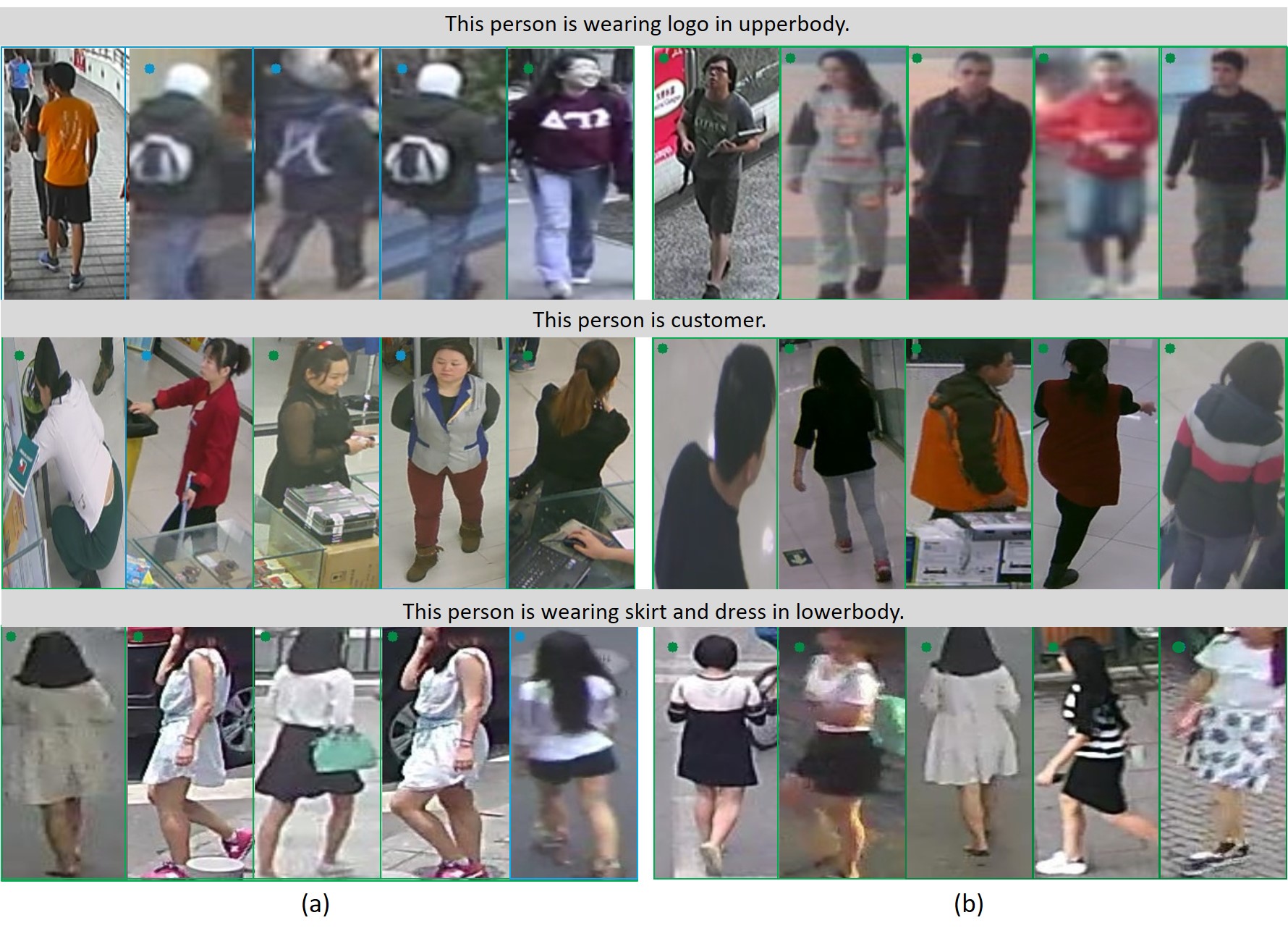}
		\caption{Text search image visualization. (a) represents the top-5 results of the CLIP method. (b) represents top-5 the results of our TEMS method. The green (blue) dot indicates the given text description is consistent (inconsistent) with the ground truth of the image. }
		\label{fig:T2I}
	\end{figure} 
	
	We also show examples of text-to-image retrieval on three datasets for models trained on the PETA dataset. Specifically, as shown in Figure~\ref{fig:T2I}, the first line is the results on the PETA dataset, the second line is the results on the RAPv1 dataset, and the third line is the results on the PA100K dataset. The second and third texts are unseen attributes. Each text corresponds to top-5 results. Figure~\ref{fig:T2I} (a) shows the results of the CLIP method, and Figure~\ref{fig:T2I} (b) shows the results of our proposed TEMS method for POAR task.
	For the PA100K and RAPv1 datasets, the TEMS model can also retrieve some correct images that belong to the unseen attributes, such as the results of ``This person is customer.''. 
	
	\begin{figure*}[t]
		\centering
		\includegraphics[width=0.75\linewidth]{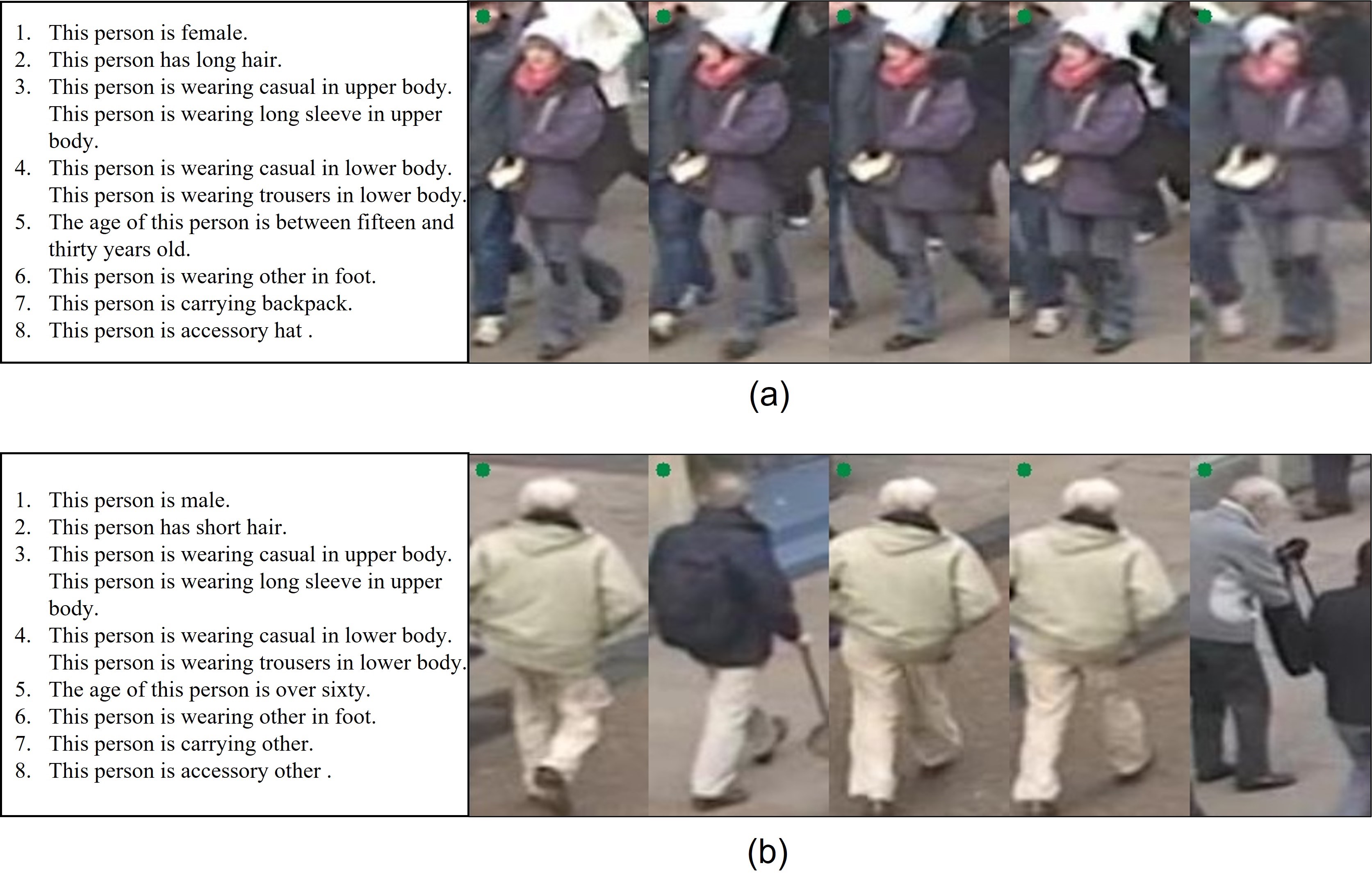}
		\caption{Text search image visualization. The green dot indicates the given text description is consistent with the ground truth of the image. }
		\label{fig:T2Iw}
	\end{figure*} 
	
	In addition, we also show examples using all the text descriptions to generate embeddings to search images on the PETA dataset. Results are shown in Figure~\ref{fig:T2Iw}. As can be seen from Figure~\ref{fig:T2Iw} (a), (b), the results of the top-5 are all correct. The proposed model retrieved correct images based on textual descriptions, which is because the description includes different attributes from different groups. 
	From these examples, we can see that using a paragraph description can also find different images of the same person, which may be helpful for person re-identification applications.
	
	\subsection{Image-to-text Retrieval Examples on Complex Scenes}
	
	Pedestrian attribute recognition in complex scenes can be effectively achieved through a two-step process involving person detection followed by attribute recognition for each individual. Here we use the WIDERAttribute dataset with 14 human attribute labels and 30 event class labels for testing. Dataset URL \url{http://mmlab.ie.cuhk.edu.hk/projects/WIDERAttribute}. 
	
\begin{figure*}[t]
		\centering
		\includegraphics[width=0.85\linewidth]{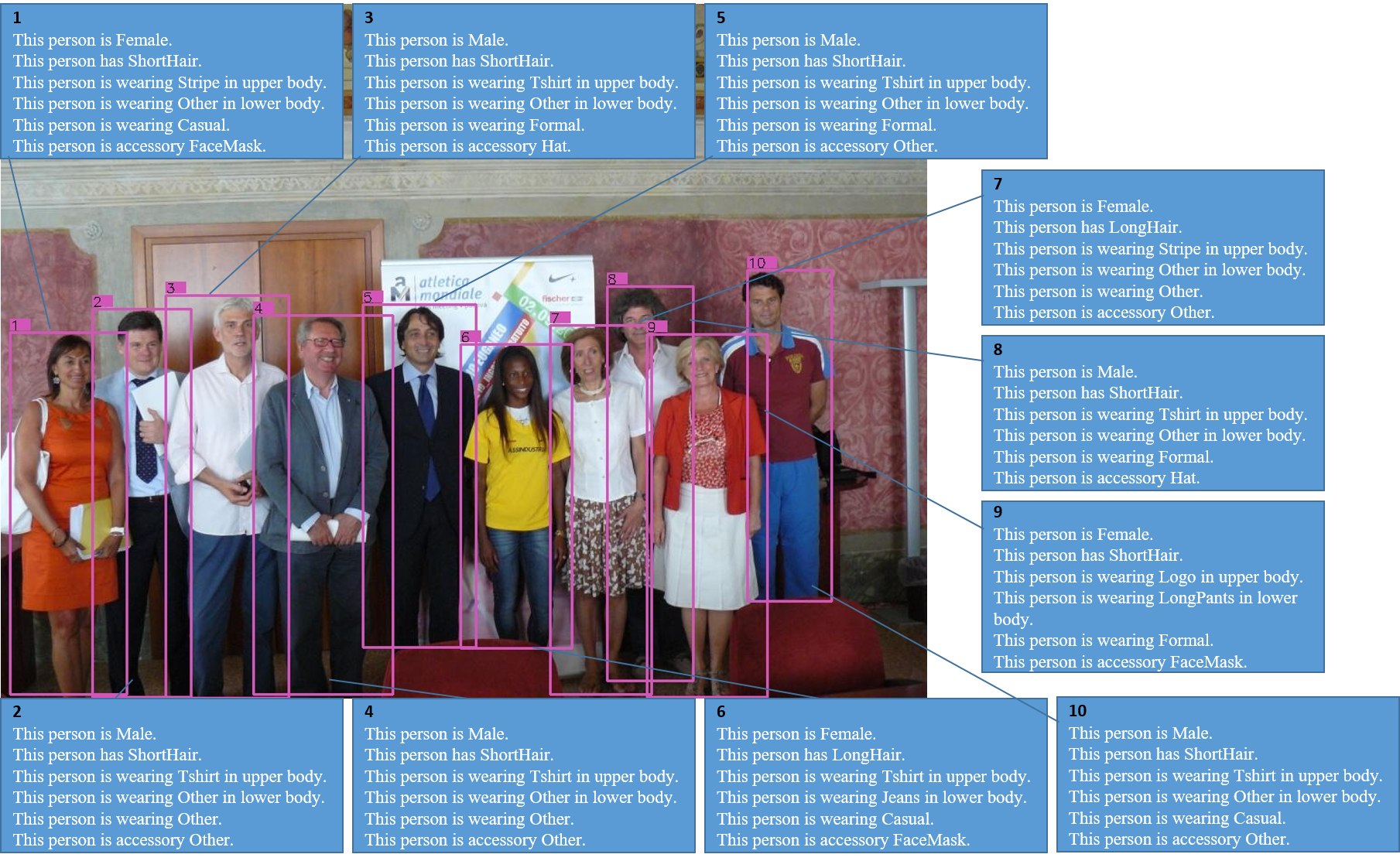}
		\caption{Pedestrian attribute recognition in complex scenes. }
		\label{fig:I2TC}
	\end{figure*}

	In Figure~\ref{fig:I2TC}, we present the outcomes of testing our model on the WIDERAttribute dataset, which was trained using the PETA dataset. The test images were entirely new and did not overlap with the training set. Moreover, these images contain multiple pedestrians, often with occlusions, making the recognition task more complex. In some instances, while identifying specific pedestrians, there could be potential influences from other pedestrians' attribute features. However, overall, our model demonstrates commendable recognition ability even in such challenging scenarios.
	
\end{document}